\documentclass[runningheads]{llncs}
\usepackage{graphicx}

\usepackage{tikz}
\usepackage{comment} 
\usepackage{amsmath,amssymb} 
\usepackage{color}

\usepackage{multirow}
\usepackage{booktabs,tabularx}
\usepackage{enumitem}
\newcommand{\minisection}[1]{\noindent{\textbf{#1}}}
\newcommand{\minienumerate}[1]{\noindent{\textbf{#1}}}

\begin{document}
\pagestyle{headings}
\mainmatter
\def\ECCVSubNumber{100}  

\title{Mask TextSpotter v3: Segmentation Proposal Network for Robust Scene Text Spotting} 

\titlerunning{Mask TextSpotter v3}
%
\author{Minghui Liao\inst{1}\thanks{Work done while an intern at Facebook}\orcidID{0000-0002-2583-4314} \and
Guan Pang\inst{2}\orcidID{0000-0002-9922-7074} \and \\
Jing Huang\inst{2}\orcidID{0000-0003-3488-3641} \and
Tal Hassner\inst{2}\orcidID{0000-0003-2275-1406} \and \\
Xiang Bai\inst{1}\thanks{Corresponding author}\orcidID{0000-0002-3449-5940}}
\authorrunning{M. Liao et al.}
%
\institute{Huazhong University of Science and Technology, China
\email{\{mhliao,xbai\}@hust.edu.cn}
\and
Facebook AI, USA \\
\email{\{gpang,jinghuang,thassner\}@fb.com}}
\maketitle

\begin{abstract}
Recent end-to-end trainable methods for scene text spotting, integrating detection and recognition, showed much progress. However, most of the current arbitrary-shape scene text spotters use region proposal networks (RPN) to produce proposals. RPN relies heavily on manually designed anchors and its proposals are represented with axis-aligned rectangles. The former presents difficulties in handling text instances of extreme aspect ratios or irregular shapes, and the latter often includes multiple neighboring instances into a single proposal, in cases of densely oriented text. To tackle these problems, we propose Mask TextSpotter v3, an end-to-end trainable scene text spotter that adopts a Segmentation Proposal Network (SPN) instead of an RPN. Our SPN is anchor-free and gives accurate representations of arbitrary-shape proposals. It is therefore superior to RPN in detecting text instances of extreme aspect ratios or irregular shapes. Furthermore, the accurate proposals produced by SPN allow masked RoI features to be used for decoupling neighboring text instances. As a result, our Mask TextSpotter v3 can handle text instances of extreme aspect ratios or irregular shapes, and its recognition accuracy won't be affected by nearby text or background noise. Specifically, we outperform state-of-the-art methods by \textbf{21.9} percent on the Rotated ICDAR 2013 dataset (rotation robustness), \textbf{5.9} percent on the Total-Text dataset (shape robustness), and achieve state-of-the-art performance on the MSRA-TD500 dataset (aspect ratio robustness). Code is available at: https://github.com/MhLiao/MaskTextSpotterV3 
\end{abstract}

\section{Introduction}\label{sec:introduction}
Reading text in the wild is of great importance, with abundant real-world applications, including Photo OCR~\cite{bissacco2013photoocr}, reading menus, and geo-location. Systems designed for this task generally consist of text detection and recognition components, where the goal of text detection is localizing the text instances with their bounding boxes whereas text recognition aims to recognize the detected text regions by converting them into a sequence of character labels. Scene text spotting/end-to-end recognition is a task that combines the two tasks, requiring both detection and recognition.

\begin{figure}[t]
\centering
\includegraphics[width=0.95\linewidth]{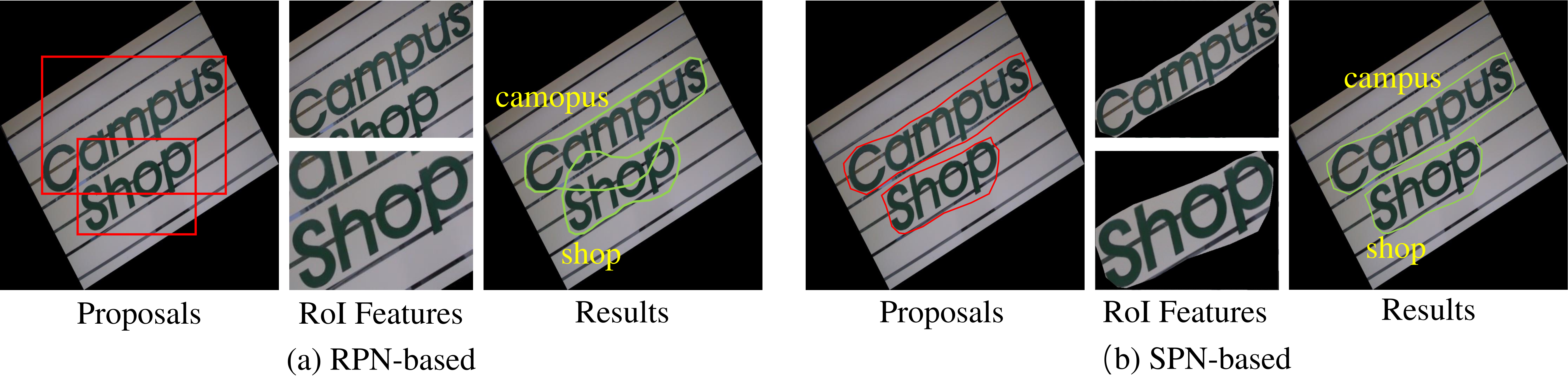}
\caption{\textbf{Comparisons between RPN and SPN.} Left: the state-of-the-art, RPN-based text spotter (Mask TextSpotter v2~\cite{liao2019mask}); Right: our SPN-based text spotter (Mask TextSpotter v3). Although RPN proposals are localized well with the axis-aligned rectangles, its RoI features contain multiple text instances, resulting in inaccurate detection/recognition. By comparison, the proposals of our SPN are more accurate, thereby producing only a single text instance for each RoI feature and leading to accurate detection/recognition results. RoIs are shown with image regions}
\label{fig:introduction}
\end{figure}

The challenges of scene text reading mainly lie in the varying orientations, extreme aspect ratios, and diverse shapes of scene text instances, which bring difficulties to both text detection and recognition. Thus, \emph{rotation robustness}, \emph{aspect ratio robustness}, and \emph{shape robustness} are necessary for accurate scene text spotters. Rotation robustness is important in scene text images, where text cannot be assumed to be well aligned with the image axes. Aspect ratio robustness is especially important for non-Latin scripts where the text is often organized in long text lines rather than words. Shape robustness is necessary for handling text of irregular shapes, which frequently appears in logos.

A recent popular trend is to perform scene text spotting by integrating both text detection and recognition into a unified model~\cite{Busta_2017_ICCV,Li_2017_ICCV}, as the two tasks are naturally closely related. Some such scene text spotters are designed to detect and recognize multi-oriented text instances, such as Liu et al.~\cite{liu2018fots} and He et al.~\cite{he2018end}. Mask TextSpotter v1~\cite{LyuLYWB18}, Qin et al.~\cite{qin2019towards}, and Mask TextSpotter v2~\cite{liao2019mask} can further handle text instances of arbitrary shapes. Mask TextSpotter series adopt Region Proposal Network (RPN)~\cite{ren2015faster} to generate proposals and extract RoI features of the proposals for detection and recognition. Qin et al.~\cite{qin2019towards} directly apply Mask R-CNN~\cite{he2017mask} for detection, which also uses RPN to produce proposals. 
These methods made great progress towards rotation robustness and shape robustness. The architectures of these methods, however, were not designed to be fully robust to rotations, aspect ratios, and shapes. Although these methods can deal with the scattered text instances of various orientations and diverse shapes, they can fail on densely oriented text instances or text lines of extreme aspect ratios due to the limitations of RPN.

The limitations of RPN mainly lie in two aspects: (1) The manually pre-designed anchors are defined using axis-aligned rectangles which cannot easily match text instances of extreme aspect ratios. (2) The generated axis-aligned rectangular proposals can contain multiple neighboring text instances when text instances are densely positioned.
As evident in Fig.~\ref{fig:introduction}, the proposals produced by Mask TextSpotter v2~\cite{liao2019mask} are overlapped with each other and its RoI features therefore include multiple neighboring text instances, causing errors for detection and recognition. As shown in Fig.~\ref{fig:introduction}, the errors can be one or several characters, which may not be embodied in the performance if a strong lexicon is given. Thus, the evaluation without lexicon or with a generic lexicon is more persuasive.

In this paper, we propose a Segmentation Proposal Network (SPN), designed to address the limitations of RPN-based methods. Our SPN is anchor-free and gives accurate polygonal representations of the proposals. Without restrictions by pre-designed anchors, SPN can handle text instances of extreme aspect ratios or irregular shapes. Its accurate proposals can then be fully utilized by applying our proposed hard RoI masking into the RoI features, which can suppress neighboring text instances or background noise. This is beneficial in cases of densely oriented or irregularly shaped texts, as shown in Fig.~\ref{fig:introduction}. Consequently, Mask TextSpotter v3 is proposed by adopting SPN into Mask TextSpotter v2.

Our experiments show that Mask TextSpotter v3 significantly improves robustness to rotations, aspect ratios, and shapes. On the Rotated ICDAR 2013 dataset where the images are rotated with various angles, our method surpasses the state-of-the-art on both detection and end-to-end recognition by more than \textbf{21.9\%}. On the Total-Text dataset~\cite{total-text} containing text instances of various shapes, our method outperforms the state-of-the-art by \textbf{5.9\%} on the end-to-end recognition task. Our method also achieves state-of-the-art performance on the MSRA-TD500 dataset~\cite{MSRA} labeled with text lines of extreme aspect ratios, as well as the ICDAR 2015 dataset that includes many low-resolution small text instances with a generic lexicon.
To summarize, our contributions are three-fold:
\begin{enumerate}[itemsep=0ex]
    \item We describe {\bf Segmentation Proposal Network (SPN)}, for an accurate representation of arbitrary-shape proposals. 
    The anchor-free SPN overcomes the limitations of RPN in handling text of extreme aspect ratios or irregular shapes, and provides more accurate proposals to improve recognition robustness.
    To our knowledge, it is the first arbitrary-shape proposal generator for end-to-end trainable text spotting.
    \item We propose {\bf hard RoI masking} to apply polygonal proposals to RoI features, effectively suppressing background noise or neighboring text instances.
    \item Our proposed {\bf Mask TextSpotter v3} significantly improves robustness to rotations, aspect ratios, and shapes, beating/achieving state-of-the-art results on several challenging scene text benchmarks.
\end{enumerate}

\section{Related work}
Current text spotting methods can be roughly classified into two categories: (1) \emph{two-stage scene text spotting} methods, whose detector and recognizer are trained separately; (2) \emph{end-to-end trainable scene text spotting} methods, which integrate detection and recognition into a unified model.

\minisection{Two-stage scene text spotting}
Two-stage scene text spotting methods use two separate networks for detection and recognition. Wang et al.~\cite{WangWCN12} tried to detect and classify characters with CNNs. 
Jaderberg et al.~\cite{jaderberg2016reading} proposed a scene text spotting method consisting of a proposal generation module, a random forest classifier to filter proposals, a CNN-based regression module for refining the proposals, and a CNN-based word classifier for recognition.
TextBoxes~\cite{liao2017textboxes} and TextBoxes++~\cite{liao2018textboxes++} combined their proposed scene text detectors with CRNN~\cite{shi2016end} and re-calculated the confidence score by integrating the detection confidence and the recognition confidence. 
Zhan et al.~\cite{Zhan_2019_ICCV} proposed to apply multi-modal spatial learning into the scene text detection and recognition system.

\begin{figure*}[ht]
    \centering
    \includegraphics[width=0.98\linewidth]{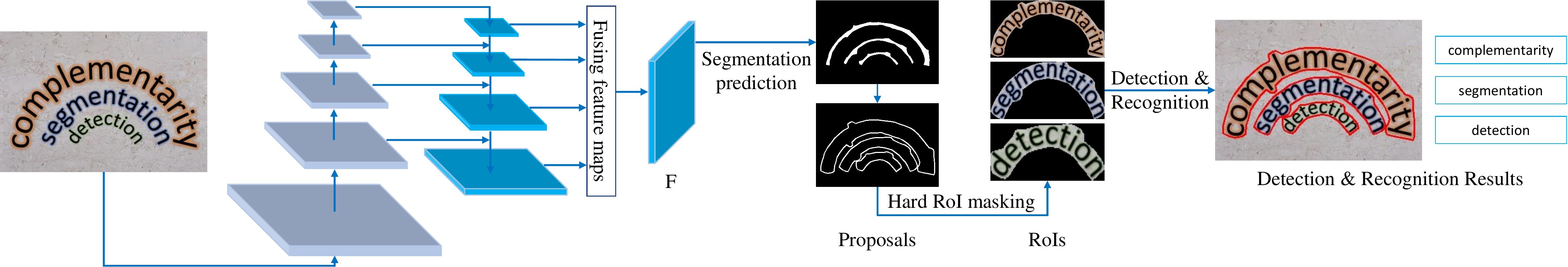}
    \caption{\textbf{Overview of Mask TextSpotter v3.} ``F'': fused feature map for segmentation. We use the original image regions to represent RoIs for better visualization}
    \label{fig:pipeline}
\end{figure*}

\minisection{End-to-end trainable scene text spotting}
Recently, end-to-end trainable scene text spotting methods have dominated this area, benefiting from the complementarity of text detection and recognition. Li et al.~\cite{Li_2017_ICCV} integrated a horizontal text detector and a sequence-to-sequence text recognizer into a unified network. Meanwhile, Bu{\v{s}}ta et al.~\cite{Busta_2017_ICCV} used a similar architecture while its detector can deal with multi-oriented text instances. After that, Liu et al.~\cite{liu2018fots} and He et al.~\cite{he2018end} further improved performance by adopting better detection and recognition methods, respectively. 

Mask TextSpotter v1~\cite{LyuLYWB18} is the first end-to-end trainable arbitrary-shape scene text spotter, consisting of a detection module based on Mask R-CNN~\cite{he2017mask} and a character segmentation module for recognition. 
Following Mask TextSpotter v1~\cite{LyuLYWB18}, several arbitrary-shape scene text spotters appeared concurrently.
Mask TextSpotter v2~\cite{liao2019mask} further extends Mask TextSpotter v1 by applying a spatial attentional module for recognition, which alleviated the problem of character-level annotations and improved the performance significantly. Qin et al.~\cite{qin2019towards} also combine a Mask R-CNN detector and an attention-based recognizer to deal with arbitrary-shape text instances. Xing et al.~\cite{xing2019charnet} propose to simultaneously detect/recognize the characters and the text instances, using the text instance detection results to group the characters. TextDragon~\cite{TextDragon} detects and recognizes text instances by grouping and decoding a series of local regions along with their centerline.

Qin et al.~\cite{qin2019towards} use the mask maps from a Mask R-CNN detector to perform RoI masking on the RoI features, which is beneficial to recognition. However, the detector that adopts RPN to produce proposals may produce inaccurate mask maps, causing further recognition errors. Different from Qin et al.~\cite{qin2019towards}, our Mask TextSpotter v3 obtains accurate proposals and applies our hard RoI masking on the RoI features for both detection and recognition modules. Thus, it can detect and recognize densely oriented/curved text instances accurately.

\minisection{Segmentation-based scene text detectors}
Zhang et al~\cite{zhang2016multi} first use FCN to obtain the salient map of the text region, then estimate the text line hypotheses by combining the salient map and character components (using MSER). Finally, another FCN predicts the centroid of each character to remove the false hypotheses. 
He et al.~\cite{he2016accurate} propose Cascaded Convolutional Text Networks (CCTN) for text center lines and text regions. PSENet~\cite{wang2019shape} adopts a progressive scale expansion algorithm to get the bounding boxes from multi-scale segmentation maps. DB~\cite{LiaoWYCB20} proposes a differentiable binarization module for a segmentation network. Comparing to the previous segmentation-based scene text detectors that adopt multiple cues or extra modules for the detection task, our method focuses on proposal generation with a segmentation network for an end-to-end scene text recognition model.

\section{Methodology}
Mask TextSpotter v3 consists of a ResNet-50~\cite{he2016deep} backbone, a Segmentation Proposal Network (SPN) for proposal generation, a Fast R-CNN module~\cite{fastrcnn} for refining proposals, a text instance segmentation module for accurate detection, a character segmentation module and a spatial attentional module for recognition. The pipeline of Mask TextSpotter v3 is illustrated in Fig.~\ref{fig:pipeline}. It provides polygonal representations for the proposals and eliminates added noise for the RoI features, thus achieving accurate detection and recognition results.

\subsection{Segmentation proposal network}
As shown in Fig.~\ref{fig:pipeline}, our proposed SPN adopts a U-Net~\cite{ronneberger2015u} structure to make it robust to scales. Unlike the FPN-based RPN~\cite{lin2017feature,ren2015faster}, which produces proposals of different scales from multiple stages, SPN generates proposals from segmentation masks, predicted from a fused feature map $F$ that concatenates feature maps of various receptive fields. $F$ is of size $\frac{H}{4} \times \frac{W}{4}$, where $H$ and $W$ are the height and width of the input image respectively. The configuration of the segmentation prediction module for $F$ is shown in the supplementary. The predicted text segmentation map $S$ is of size $1 \times H \times W$, whose values are in the range of $[0, 1]$.

\begin{figure}[ht]
\centering
\includegraphics[width=0.7\linewidth]{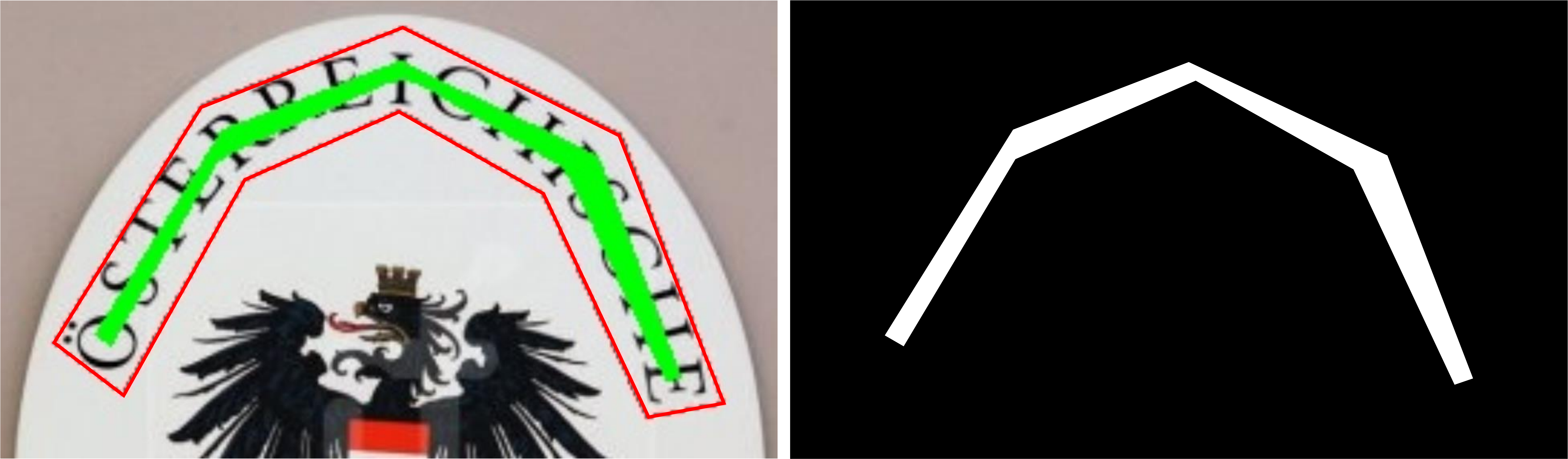}
\caption{\textbf{Illustration of the segmentation label generation.} Left: Red and green polygons are the original annotation and the shrunk region respectively. Right: segmentation label; black and white indicate the values of $0$ and $1$ respectively}
\label{fig:label_generation}
\end{figure}

\minisection{Segmentation label generation}\label{sec:label_generation}
To separate the neighboring text instances, it is common for segmentation-based scene text detectors to shrink the text regions~\cite{zhou2017east,wang2019shape}. Inspired by Wang et al.~\cite{wang2019shape} and DB~\cite{LiaoWYCB20}, we adopt the Vatti clipping algorithm~\cite{vati1992generic} to shrink the text regions by clipping $d$ pixels. The offset pixels $d$ can be determined as 
$d = A(1-r^{2})/{L}$,
where $A$ and $L$ are the area and perimeter of the polygon that represents the text region, and $r$ is the shrink ratio, which we empirically set to $0.4$. An example of the label generation is shown in Fig.~\ref{fig:label_generation}.

\minisection{Proposal generation} \label{sec:propsal_generation}
Given a text segmentation map, $S$, whose values are in the range of $[0, 1]$, we first binarize $S$ into a binary map $B$:
\begin{equation}
    B_{i, j} = 
    \begin{cases}
        1& \text{if } S_{i, j} >= t, \\
        0&  \text{otherwise.}
    \end{cases}
    \label{eq:binarization}
\end{equation}
Here, $i$ and $j$ are the indices of the segmentation or binary map and $t$ is set to $0.5$. Note that $B$ is of the same size as $S$ and the input image.

We then group the connected regions in the binary map $B$. These connected regions can be considered as shrunk text regions since the text segmentation labels are shrunk, as described above. Thus, we dilate them by un-clipping $\hat{d}$ pixels using the Vatti clipping algorithm, where $\hat{d}$ is calculated as 
$\hat{d} = \hat{A} \times \hat{r}/{\hat{L}}$.
Here, $\hat{A}$ and $\hat{L}$ are the area and perimeter of the predicted shrunk text regions. $\hat{r}$ is set to $3.0$ according to the value of the shrink ratio $r$.

As explained above, the proposals produced by SPN can be accurately represented as polygons, which are the contours of text regions. Thus, SPN generates suitable proposals for text lines with extreme aspect ratios and densely oriented/irregularly shaped text instances. 

\subsection{Hard RoI masking} \label{sec:roi_masking}
Since the custom RoI Align operator only supports the axis-aligned rectangular bounding boxes, we use the minimum, axis-aligned, rectangular bounding boxes of the polygon proposals to generate the RoI features to keep the RoI Align operator simple.

Qin et al.~\cite{qin2019towards} proposed RoI masking which multiplies the mask probability map and the RoI feature, where the mask probability map is generated by a Mask R-CNN detection module. However, the mask probability maps may be inaccurate since they are predicted by the proposals from RPN. For example, it may contain multiple neighboring text instances for densely oriented text. In our case, accurate polygonal representations are designed for the proposals, thus we can directly apply the proposals to the RoI features through our proposed hard RoI masking.

Hard RoI masking multiplies binary polygon masks with the RoI features to suppress background noise or neighboring text instances, where a polygon mask $M$ indicates an axis-aligned rectangular binary map with all $1$ values in the polygon region and all $0$ values outside the polygon region. Assuming that $R_0$ is the RoI feature and $M$ is the polygon mask, which is of size $32 \times 32$, the masked RoI feature $R$ can be calculated as $R = R_0 * M$,
where $*$ indicates element-wise multiplication. $M$ can be easily generated by filling the polygon proposal region with $1$ while setting the values outside the polygon to $0$. We report an ablation study on the hard RoI masking in Sec.~\ref{sec:ablation}, where we compare the proposed hard RoI masking with other operators including the RoI masking in Qin et al.~\cite{qin2019towards}.

After applying hard RoI masking, the background regions or neighboring text instances are suppressed in our masked RoI features, which significantly reduce the difficulties and errors in the detection and recognition modules.

\subsection{Detection and recognition}
We follow the main design of its text detection and recognition modules of Mask TextSpotter v2~\cite{liao2019mask} for the following reasons: (1) Mask TextSpotter v2 is the current state-of-the-art with competitive detection and recognition modules. (2) Since Mask TextSpotter v2 is a representative method in the RPN-based scene text spotters, we can fairly compare our method with it to verify the effectiveness and robustness of our proposed SPN.

For detection, the masked RoI features generated by the hard RoI masking are fed into the Fast R-CNN module for further refining the localizations and the text instance segmentation module for precise segmentation. The character segmentation module and spatial attentional module are adopted for recognition.

\subsection{Optimization}
The loss function $L$ is defined as below:
\begin{equation}
    L = L_s + \alpha_1 L_{rcnn} + \alpha_2 L_{mask}.
\end{equation}
$L_{rcnn}$ and $L_{mask}$ are defined in Fast R-CNN~\cite{fastrcnn} and Mask TextSpotter v2~\cite{liao2019mask} respectively. $L_{mask}$ consists of a text instance segmentation loss, a character segmentation loss, and a spatial attentional decoder loss. $L_s$ indicates the SPN loss. Finally, following Mask TextSpotter v2~\cite{liao2019mask}, we set $\alpha_1$ and $\alpha_2$ to~1.0.

We adopt dice loss~\cite{milletari2016v} for SPN. Assuming that $S$ and $G$ are the segmentation map and the target map, the segmentation loss $L_s$ can be calculated as:
\begin{gather}
        I = \sum (S * G); \hspace{4mm}  
        U = \sum S + \sum G; \hspace{4mm}
        L_s = 1 - \frac{2.0 \times I}{U}, 
\end{gather}
where $I$ and $U$ indicate the intersection and union of the two maps, and $*$ represents element-wise multiplication.

\section{Experiments}
We evaluate our method, testing robustness to four types of variations: rotations, aspect ratios, shapes, and small text instances, on different standard scene text benchmarks.
We further provide an ablation study of our hard RoI masking.

\subsection{Datasets}
\minienumerate{SynthText}~\cite{gupta2016synthetic} is a synthetic dataset containing 800k text images. It provides annotations for word/character bounding boxes and text sequences.

\minisection{Rotated ICDAR 2013 dataset (RoIC13)} is generated from the ICDAR 2013 dataset~\cite{karatzas2013icdar}, whose images are focused around the text content of interest. The text instances are in the horizontal direction and labeled by axis-aligned rectangular boxes. Character-level segmentation annotations are given and so we can get character-level bounding boxes. The dataset contains 229 training and 233 testing images. 
To test rotation robustness, we create the Rotated ICDAR 2013 dataset by rotating the images and annotations in the test set of the ICDAR 2013 benchmark with some specific angles, including $15^\circ$, $30^\circ$, $45^\circ$, $60^\circ$, $75^\circ$, and $90^\circ$. Since all text instances in the ICDAR 2013 dataset are horizontally oriented, we can easily control the orientations of the text instances and find the relations between performances and text orientations. We use the evaluation protocols in the ICDAR 2015 dataset, because the ones in ICDAR 2013 only support axis-aligned bounding boxes.

\minisection{MSRA-TD500 dataset}~\cite{MSRA} is a multi-language scene text detection benchmark that contains English and Chinese text, including 300 training images and 200 testing images. Text instances are annotated in the text-line level, thus there are many text instances of extreme aspect ratios. This dataset does not contain recognition annotations.

\minisection{Total-Text dataset}~\cite{total-text,CK2019} includes 1,255 training and 300 testing images. It offers text instances of various shapes, including horizontal, oriented, and curved shapes, which are annotated with polygonal bounding boxes and transcriptions. Note that although character-level annotations are provided in the Total-Text dataset, we do not use them for fair comparisons with previous methods\cite{lyu2018multi,liao2019mask}.

\minisection{ICDAR 2015 dataset (IC15)}~\cite{karatzas2015icdar} consists of 1,000 training images and 500 testing images, which are annotated with quadrilateral bounding boxes. Most of the images are of low resolution and contain small text instances.

\subsection{Implementation details}
For a fair comparison with Mask TextSpotter v2~\cite{liao2019mask}, we use the same training data and training settings described below. Data augmentation follows the official implementation of Mask TextSpotter v2~\footnote{\label{mtscode}https://github.com/MhLiao/MaskTextSpotter}, including multi-scale training and pixel-level augmentations. Since our proposed SPN can deal with text instances of arbitrary shapes and orientations without conflicts, we adopt a more radical rotation data augmentation. The input images are randomly rotated with an angle range of $[-90^\circ, 90^\circ]$ while the original Mask TextSpotter v2 uses an angle range of $[-30^\circ, 30^\circ]$. Note that the Mask TextSpotter v2 is trained with the same rotation augmentation as ours for the experiments on the RoIC13 dataset.

The model is optimized using SGD with a weight decay of $0.001$ and a momentum of $0.9$. It is first pre-trained with SynthText and then fine-tuned with a mixture of SynthText, the ICDAR 2013 dataset, the ICDAR 2015 dataset, the SCUT dataset~\cite{zhong2016deeptext}, and the Total-Text dataset for 250k iterations. The sampling ratio among these datasets is set to $2:2:2:1:1$ for each mini-batch of eight.

During pre-training, the learning rate is initialized with $0.01$ and then decreased to a tenth at 100k iterations and 200k iterations respectively. During fine-tuning, we adopt the same training scheme while using $0.001$ as the initial learning rate. We choose the model weights of 250k iterations for both pre-training and fine-tuning.
In the inference period, the short sides of the input images are resized to $1000$ on the RoIC13 dataset and the Total-Text dataset, $1440$ on the IC15 dataset, keeping the aspect ratios.

\begin{figure}[ht]
    \centering
    \includegraphics[width=0.7\linewidth]{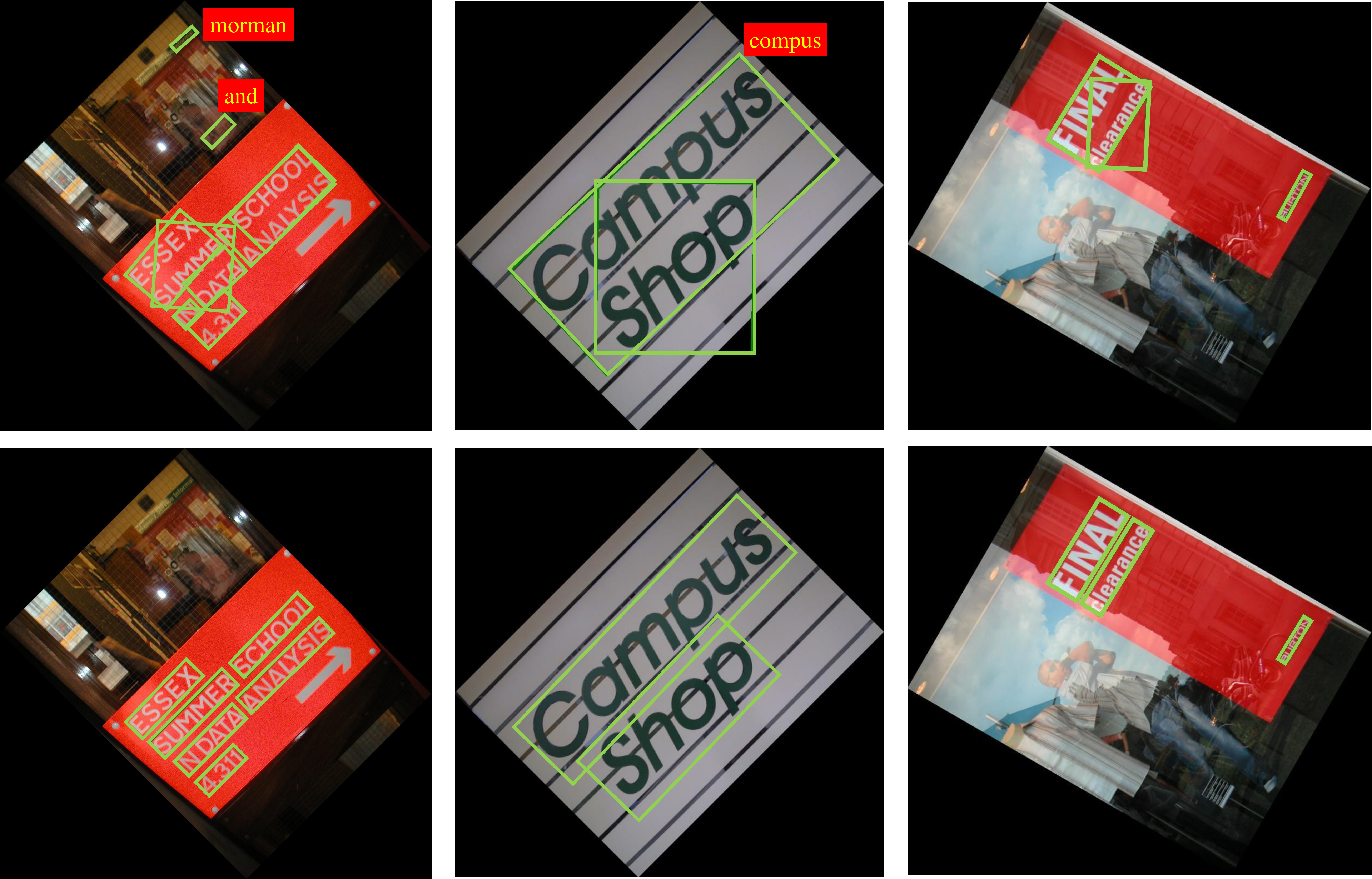}
    \caption{\textbf{Qualitative results on the RoIC13 dataset.} Top: Mask TextSpotter v2; Bottom: Mask TextSpotter v3. More results in the supplementary}
    \label{fig:rotation-visu}
\end{figure}

\begin{figure}[ht]
    \centering
    \includegraphics[width=0.7\linewidth]{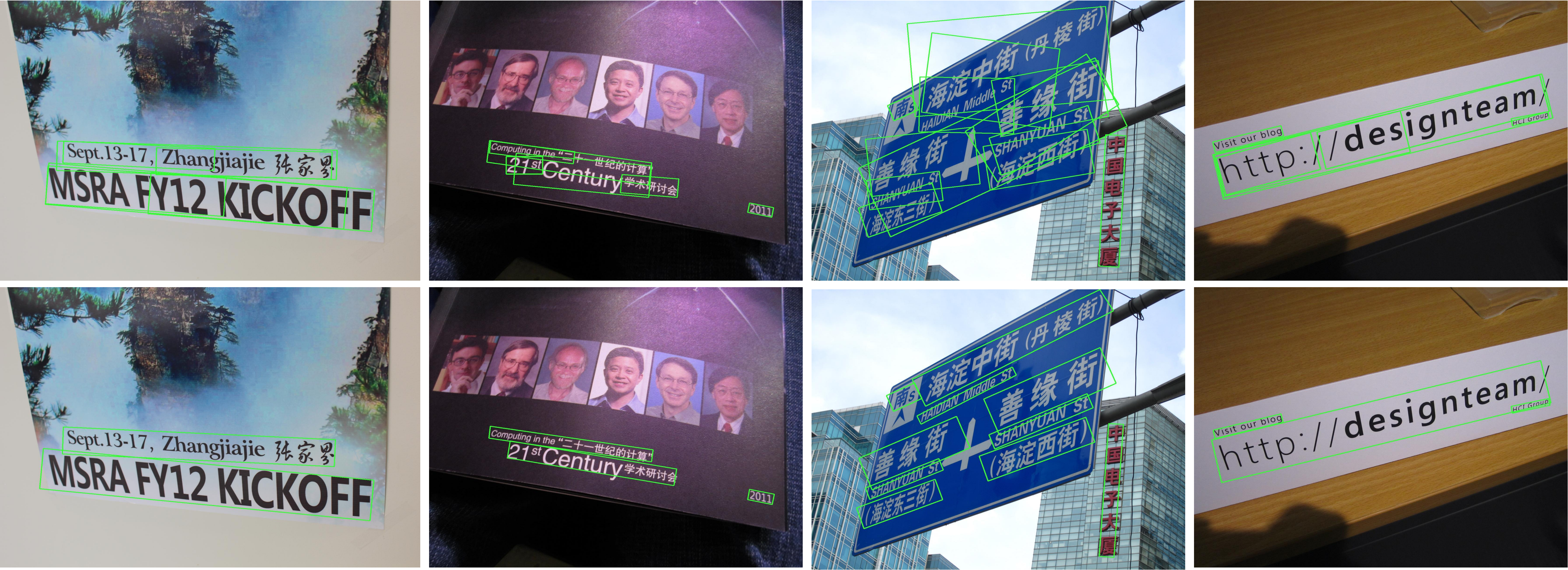}
    \caption{\textbf{Qualitative results on the MSRA-TD500 dataset.} Top: Mask TextSpotter v2; Bottom: Mask TextSpotter v3}
    \label{fig:td500_visu}
\end{figure}

\begin{figure}[ht]
    \centering
    \includegraphics[width=0.7\linewidth]{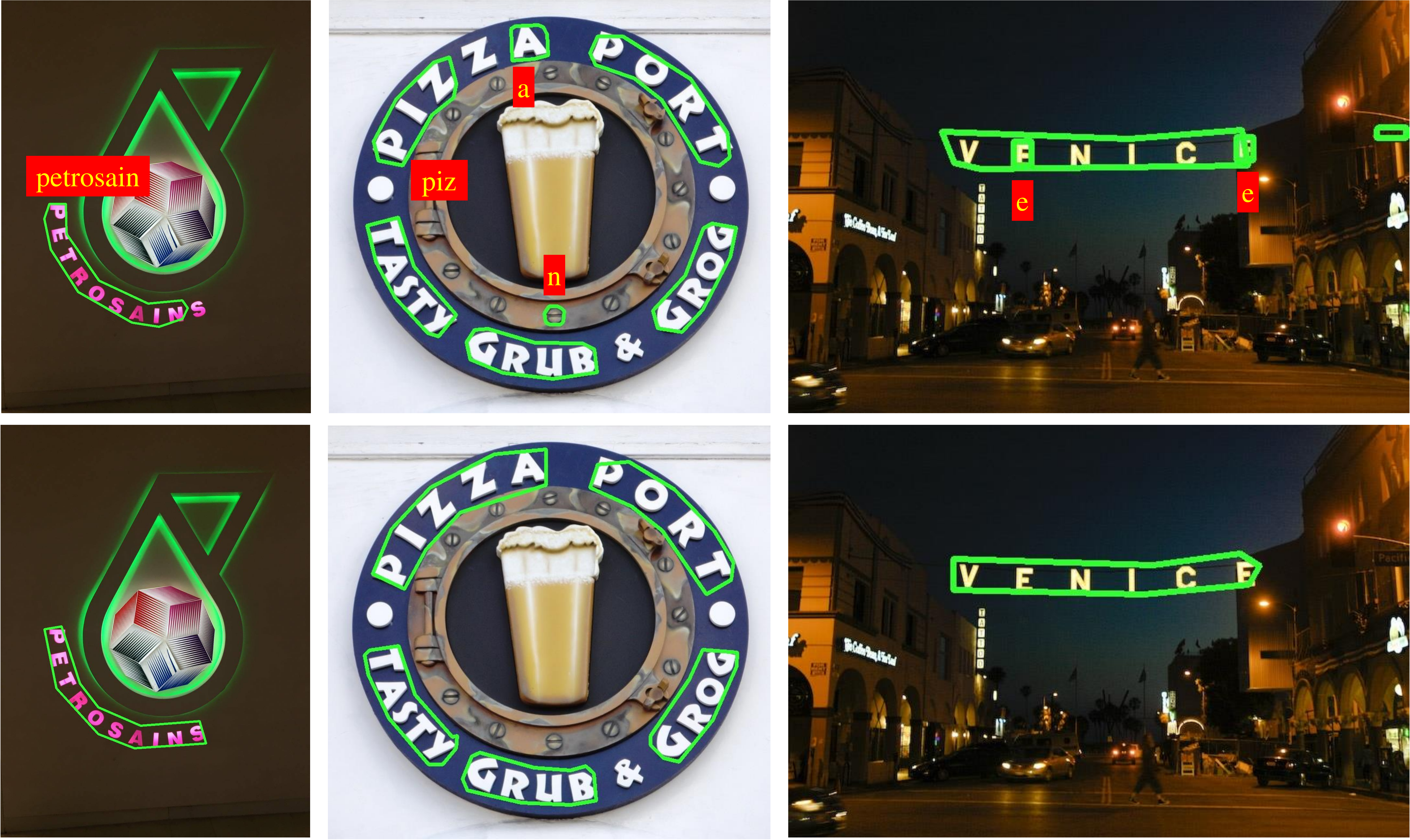}
    \caption{\textbf{Qualitative results on the Total-Text dataset.} Top: Mask TextSpotter v2; Bottom: Mask TextSpotter v3. The yellow text with red background are some inaccurate recognition results. Only inaccurate recognition results are visualized}
    \label{fig:shape_visu}
\end{figure}

\subsection{Rotation robustness}
We test for rotation robustness by conducting experiments on the RoIC13 dataset. We compare the proposed Mask TextSpotter v3 with two state-of-the-art methods Mask TextSpotter v2~\textsuperscript{\ref{mtscode}} and CharNet~\footnote{https://github.com/MalongTech/research-charnet}, with their official implementations. For a fair comparison, Mask TextSpotter v2 is trained with the same data and data augmentation as ours. Some qualitative comparisons on the RoIC13 dataset are shown in Fig.~\ref{fig:rotation-visu}. We can see that Mask TextSpotter v2 fails on detecting and recognizing the densely oriented text instances while Mask TextSpotter v3 can successfully handle such cases.

We use the pre-trained model with a large backbone (Hourglass-88~\cite{hourglass}) for CharNet since the official implementation does not provide the ResNet-50 backbone. Note that the official pre-trained model of CharNet is trained with different training data. Thus, it is not suitable to directly compare the performance with Mask TextSpotter v3. However, we can observe the performance variations under different rotation angles. The detection and end-to-end recognition performance of CharNet drop dramatically when the rotation angle is large.

\begin{figure}[ht]
    \centering
    \begin{tabular}{l}
    \includegraphics[width=0.4\linewidth]{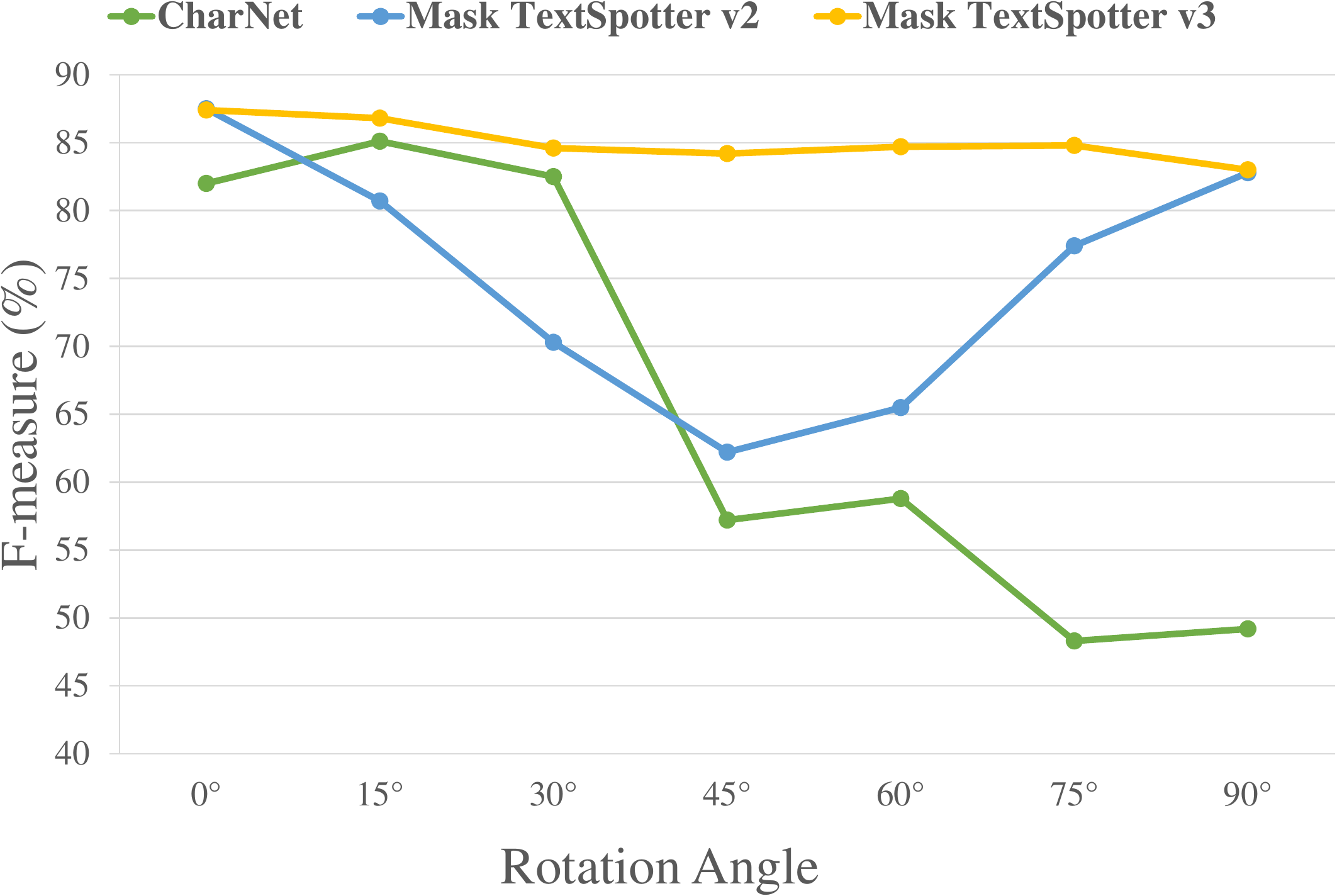}
    \label{fig:rotation-detection}
    \end{tabular}{}
    \begin{tabular}{r}
    \includegraphics[width=0.4\linewidth]{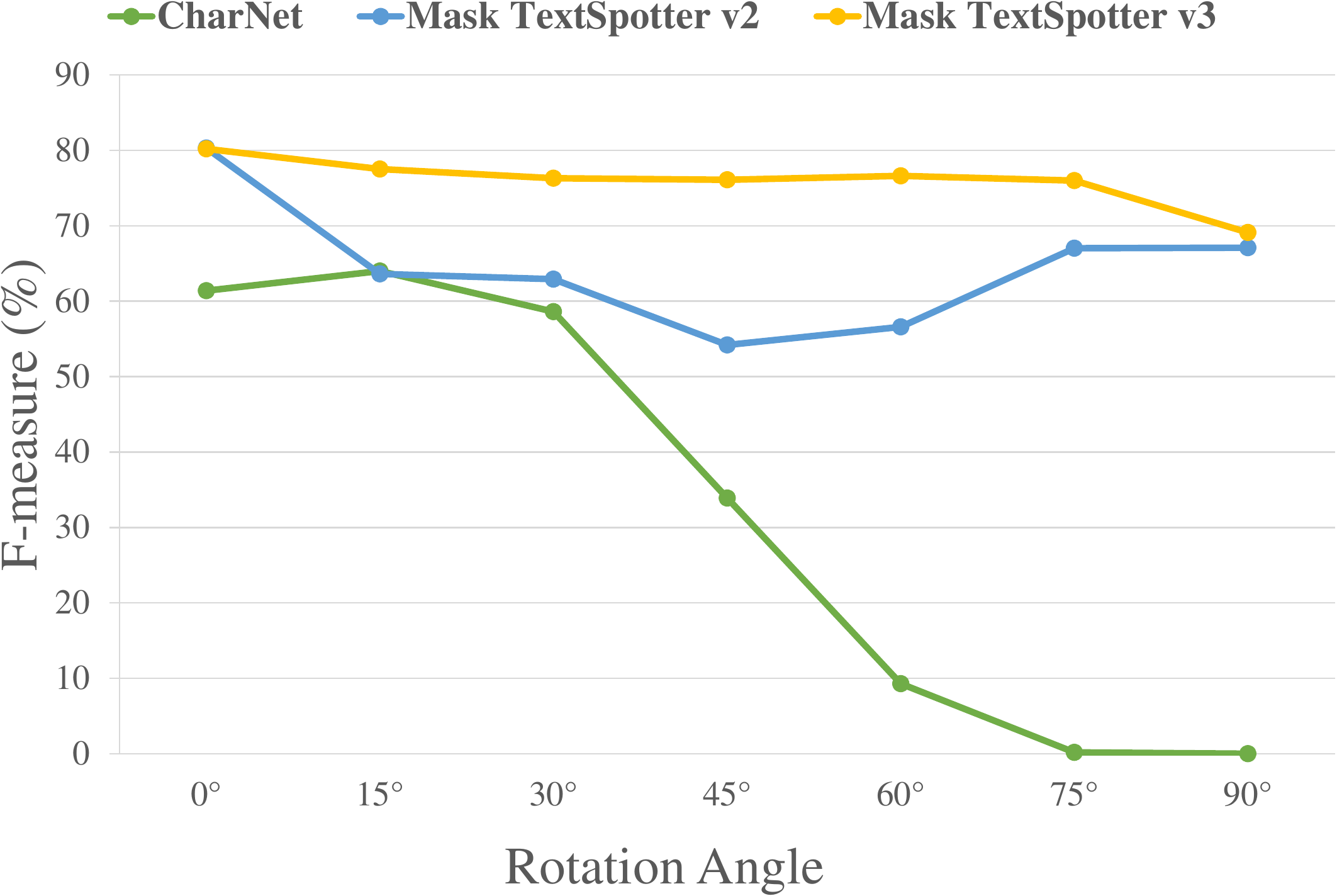}
    \label{fig:rotation-recognition}
    \end{tabular}{}
    \caption{\textbf{Detection (left) and end-to-end recognition (right) results on the RoIC13 dataset with different rotation angles.} The recognition results are evaluated without lexicon. Mask TextSpotter v2 is trained with the same rotation augmentation as Mask TextSpotter v3. CharNet is tested with the official released pre-trained model, with a backbone of Hourglass-88~\cite{hourglass}}
\end{figure}

\begin{table*}[ht]
    \setlength{\tabcolsep}{2.5pt}
    \centering
    \caption{\textbf{Quantitative results on the RoIC13 dataset.} The evaluation protocol is the same as the one in the IC15 dataset. The end-to-end recognition task is evaluated without lexicon. *CharNet is tested with the officially released pre-trained model; Mask TextSpotter v2 (MTS v2) is trained with the same rotation augmentation as Mask TextSpotter v3 (MTS v3).  ``P'', ``R'', and ``F'' indicate precision, recall and F-measure. ``E2E'' is short for end-to-end recognition. More results are in the supplementary}
    \begin{tabularx}{1.0\textwidth}{lc*{5}ccc*{4}c}
    \toprule
    \multirow{3}{*}{Method} & \multicolumn{6}{@{}c@{}}{\begin{tabular}[c]{@{}c@{}}RoIC13 dataset\\ (Rotation Angle: $45^\circ$)\end{tabular}}                                 & \multicolumn{6}{@{}c@{}}{\begin{tabular}[c]{@{}c@{}}RoIC13 dataset\\ (Rotation Angle: $60^\circ$)\end{tabular}}                                 \\ \cmidrule{2-13} 
    & \multicolumn{3}{@{}c@{}}{Detection}                 
    & \multicolumn{3}{@{}c@{}}{\begin{tabular}[c]{@{}c@{}}E2E\end{tabular}} & \multicolumn{3}{@{}c@{}}{Detection}                         & \multicolumn{3}{@{}c@{}}{\begin{tabular}[c]{@{}c@{}}E2E\end{tabular}} \\ \cmidrule{2-13} 
                                & P                & R                & F                & P                           & R                           & F                         & P                & R                & F                & P                           & R                           & F                         \\
    \midrule 
    CharNet*~\cite{xing2019charnet}     & 57.8             & 56.6             & 57.2             & 34.2                        & 33.5                       & 33.9                      & 65.5             & 53.3             & 58.8             & 10.3                        & 8.4                        & 9.3                     \\ 
    MTS v2*~\cite{liao2019mask}     & 64.8             & 59.9             & 62.2             & 66.4                        & 45.8                        & 54.2                      & 70.5             & 61.2             & 65.5             & 68.2                        & 48.3                        & 56.6                      \\ \hline
    \textbf{MTS v3}     & \textbf{91.6}    & \textbf{77.9}    & \textbf{84.2}    & \textbf{88.5}               & \textbf{66.8}               & \textbf{76.1}             & \textbf{90.7}    & \textbf{79.4}    & \textbf{84.7}    & \textbf{88.5}               & \textbf{67.6}               & \textbf{76.6}             \\ 
    \bottomrule
    \end{tabularx}
    \label{tab:rotated_ic13}
\end{table*}

\minisection{Detection task}
As shown in Fig.~\ref{fig:rotation-detection}, the detection performance of Mask TextSpotter v2 drops dramatically when the rotation angles are $30^\circ$, $45^\circ$, and $60^\circ$. In contrast, the detection results of Mask TextSpotter v3 are much more stable with various rotation angles. The maximum performance gap between Mask TextSpotter v3 and Mask TextSpotter v2 occurs when the rotation angle is $45^\circ$. As shown in Tab.~\ref{tab:rotated_ic13}, Mask TextSpotter v3 outperforms Mask TextSpotter v2 by \textbf{26.8} percent, \textbf{18.0} percent, and \textbf{22.0} percent in terms of Precision, Recall, and F-measure, with a rotation angle of $45^\circ$. Note that it is reasonable that the two methods achieve almost the same results with $0^\circ$ and $90^\circ$, since $0^\circ$ indicates without rotation and the bounding boxes are also in the shape of axis-aligned rectangles when the rotation angle is $90^\circ$.

\minisection{End-to-end recognition task}
The trend of the end-to-end recognition results is similar to the detection results, as shown in Fig.~\ref{fig:rotation-recognition}. The performance gaps between Mask TextSpotter v2 and Mask TextSpotter v3 are especially large when the rotation angles are $30^\circ$, $45^\circ$, and $60^\circ$. Mask TextSpotter v3 surpasses Mask TextSpotter v2 by more than \textbf{19.2} percent in terms of F-measure with the rotation angle of $45^\circ$ and $60^\circ$. The detailed results of $45^\circ$ rotation angle are listed in Tab.~\ref{tab:rotated_ic13}, where Mask TextSpotter v3 achieves \textbf{22.1}, \textbf{21.0}, and \textbf{21.9} performance gain compared to the previous state-of-the-art method Mask TextSpotter v2.

The qualitative and quantitative results on the detection task and end-to-end recognition task prove the rotation robustness of Mask TextSpotter v3. The reason is the RPN used in Mask TextSpotter v2 would result in errors in both detection and recognition when dealing with densely oriented text instances. In contrast, the proposed SPN can generate accurate proposals and exclude the neighboring text instances by hard RoI masking in such cases. More qualitative and quantitative results are provided in the supplementary.

\begin{table}[ht]
    \setlength{\tabcolsep}{11.0pt}
    \centering
    \caption{Quantitative detection results on the MSRA-TD500 dataset}
    \begin{tabularx}{0.72\linewidth}{lccc}
    \toprule
    Method        & P             & R             & F                   \\ 
    \midrule
    He et al.~\cite{he2016text}        & 71            & 61            & 69              \\ 
    DeepReg~\cite{deepdirect}     & 77            & 70            & 74                \\ 
    RRD~\cite{liao2018rotation}          & 87            & 73            & 79                   \\ 
    PixelLink~\cite{deng2018pixellink}    & 83.0            & 73.2          & 77.8                   \\ 
    Xue et al.~\cite{xue2018accurate}        & 83.0          & 77.4          & 80.1                \\ 
    CRAFT~\cite{craft}        & 88.2          & 78.2          & 82.9                \\ 
    Tian et al.~\cite{tian2019learning}       & 84.2          & \textbf{81.7}          & 82.9               \\ 
    MSR~\cite{tian2019learning}       & 87.4          & 76.7          & 81.7              \\ 
    DB (without DCN)~\cite{LiaoWYCB20}       & 86.6            & 77.7          & 81.9              \\ 
    Mask TextSpotter v2~\cite{liao2019mask} & 80.8         & 68.6         & 74.2           \\ 
    \hline
    \textbf{Mask TextSpotter v3} & \textbf{90.7} & 77.5 & \textbf{83.5}           \\ 
    \bottomrule
    \end{tabularx}
    \label{tab:td500}
\end{table}

\subsection{Aspect ratio robustness}
Aspect ratio robustness is verified by our experimental results on the MSRA-TD500 dataset, which contains many text lines of extreme aspect ratios. Since there are no recognition annotations, we disable our recognition module and evaluate only on the detection task. Our qualitative and quantitative results are shown in Fig.~\ref{fig:td500_visu} and Tab.~\ref{tab:td500}.

Although Mask TextSpotter v2 is the existing state-of-the-art, end-to-end recognition method, it fails to detect long text lines due to the limitation of RPN. Compared with Mask TextSpotter v2, Mask TextSpotter v3 achieves a \textbf{9.3\%} performance gain, which proves its superiority in handling text lines of extreme aspect ratios. Moreover, Mask TextSpotter v3 even outperforms state-of-the-art methods designed for text line detection~\cite{mcn,craft,tian2019learning}, further showing its robustness to aspect ratio variations.

\subsection{Shape robustness}
Robustness to shape variations is evaluated with end-to-end recognition performance on the Total-Text dataset, which contains text instances of various shapes, including horizontal, oriented, and curved shapes. Some qualitative results are shown in Fig.~\ref{fig:shape_visu}, where we can see that our method obtains more accurate detection and recognition results compared with Mask TextSpotter v2, especially on text instances with irregular shapes or with large spaces between neighboring characters. The quantitative results listed in Tab.~\ref{tab:total-text} show that our method outperforms Mask TextSpotter v2 by \textbf{5.9\%} in terms of F-measure when no lexicon is provided. Both the qualitative and quantitative results demonstrate the superior robustness to shape variations offered by our method.

\begin{table}[ht]
    \setlength{\tabcolsep}{13.0pt}
    \centering
    \caption{\textbf{Quantitative end-to-end recognition results on the Total-Text dataset.} ``None'' means recognition without any lexicon. ``Full'' lexicon contains all words in the test set. The values in the table are the F-measure. The evaluation protocols are the same as those in Mask TextSpotter v2
    }
    \begin{tabularx}{0.65\linewidth}{lcc}
    \toprule
    Method    & None               & Full       \\ 
    \midrule
    Mask TextSpotter v1~\cite{LyuLYWB18}          & 52.9               & 71.8               \\ 
    CharNet~\cite{xing2019charnet} Hourglass-57               & 63.6               & -                  \\ 
    Qin et al.~\cite{qin2019towards} Inc-Res           & 63.9               & -                  \\ 
    Boundary TextSpotter~\cite{WangLZYBXHW020}          & 65.0               & 76.1                  \\ 
    ABCNet~\cite{abcnet}          & 64.2               & 75.7                  \\ 
    Mask TextSpotter v2~\cite{liao2019mask}          & 65.3               & 77.4               \\ 
    \hline
    \textbf{Mask TextSpotter v3} & \textbf{71.2}      & \textbf{78.4}      \\
    \bottomrule 
    \end{tabularx}
    \label{tab:total-text}
\end{table}

\subsection{Small text instance robustness}
The challenges in the IC15 dataset mainly lie in the low-resolution and small text instances. As shown in Tab.~\ref{tab_icdar2015}, Mask TextSpotter v3 outperforms Mask TextSpotter v2 on all tasks with different lexicons, demonstrating the superiority of our method on handling small text instances in low-resolution images.

Although TextDragon~\cite{TextDragon} achieves better results on some tasks with the strong/weak lexicon, our method outperforms it by large margins, $7.1\%$ and $9.0\%$, with the generic lexicon. We argue that there are no such strong/weak lexicons with only 100/1000+ words in most real-world applications, thus performance with a generic lexicon of 90k words is more meaningful and more challenging. Regardless, the reason for the different behaviors with different lexicons is that the attention-based recognizer in our method can learn the language knowledge while the CTC-based recognizer in TextDragon is more independent for the character prediction. Mask TextSpotter v3 relies less on the correction of the strong lexicon, which is also one of the advantages.

\begin{table*}[ht]
    \setlength{\tabcolsep}{5.0pt}
    \centering
    \caption{\textbf{Quantitative results on the IC15 dataset} in terms of F-measure. ``S'', ``W'' and ``G'' mean recognition with strong, weak, and generic lexicon respectively. The values in the bracket (such as $1,600$ and $1,400$) indicate the short side of the input images. Note that in most real-world applications there are no such strong/weak lexicons with only 100/1000+ words. Thus, performance with the generic lexicon of 90k words is more meaningful
    }
    \begin{tabularx}{1.0\textwidth}{lc*{6}c}
    \toprule
    \multirow{2}{*}{Method} & \multicolumn{3}{@{}c@{}}{Word Spotting} & \multicolumn{3}{@{}c@{}}{E2E Recognition} & \multirow{2}{*}{FPS}\\
    \cmidrule{2-7} 
     & S & W & G & S & W & G & \\
    \midrule
     TextBoxes++~\cite{liao2018textboxes++}  &76.5 &69.0 &54.4 &73.3 &65.9 &51.9 &-
     \\
      
     He~\emph{et al.}~\cite{he2018end}  &85.0 &80.0 &65.0 &82.0 &77.0 &63.0 &-
     \\
     
    Mask TextSpotter v1~\cite{LyuLYWB18} (1600)  &79.3  &74.5  &64.2  &79.3  &73.0  &62.4  &2.6   \\
    
    TextDragon~\cite{TextDragon}  &\textbf{86.2}  &\textbf{81.6}  &68.0  &82.5  &\textbf{78.3}  &65.2  &2.6   \\
    
    CharNet~\cite{xing2019charnet} R-50 &-  &-  &-  &80.1  &74.5  &62.2  &-   \\
    
    Boundary TextSpotter~\cite{WangLZYBXHW020} &-  &-  &-  &79.7  &75.2  &64.1  &-   \\
    
    Mask TextSpotter v2~\cite{liao2019mask} (1600)  &82.4  &78.1  &73.6  &83.0  &77.7  &73.5  &2.0   \\
    \hline
    \textbf{Mask TextSpotter v3} (1440)  &83.1  &79.1  &\textbf{75.1}  &\textbf{83.3}  &78.1  &\textbf{74.2}  &2.5   \\
    \bottomrule
    \end{tabularx}
    \label{tab_icdar2015}
\end{table*}

\subsection{Ablation study} \label{sec:ablation}
It is important to apply polygon-based proposals to the RoI features. There are two attributions for such an operator: ``direct/indirect'' and ``soft/hard''. ``direct/indirect'' means using the segmentation/binary map directly or through additional layers; ``soft/hard'' indicates a soft probability mask map whose values are from $[0, 1]$ or a binary polygon mask map whose values are $0$ or $1$. We conduct experiments on four types of combinations and the results show that our proposed hard RoI masking (Direct-hard) is simple yet achieves the best performance. Results and discussions are in the supplementary.

\subsection{Limitations}
Although Mask TextSpotter v3 is far more robust to rotated text variations than the existing state-of-the-art scene text spotters, it still suffers minor performance disturbance with some extreme rotation angles, e.g. $90^\circ$, as shown in Fig.~\ref{fig:rotation-recognition}, since it is hard for the recognizer to judge the direction of the text sequence. In the future, we plan to make the recognizer more robust to such rotations.

\section{Conclusion}
We propose Mask TextSpotter v3, an end-to-end trainable arbitrary-shape scene text spotter. It introduces SPN to generate proposals, represented with accurate polygons. Thanks to the more accurate proposals, Mask TextSpotter v3 is much more robust on detecting and recognizing text instances with rotations or irregular shapes than previous arbitrary-shape scene text spotters that use RPN for proposal generation. Our experiment results on the Rotated ICDAR 2013 dataset with different rotation angles, the MSRA-TD500 dataset with long text lines, and the Total-Text dataset with various text shapes demonstrate the robustness to rotations, aspect ratios, and shape variations of Mask TextSpotter v3. Moreover, results on the IC15 dataset show that the proposed Mask TextSpotter v3 is also robust in detecting and recognizing small text instances. We hope the proposed SPN could extend the application of OCR to other challenging domains~\cite{hassner2012computation} and offer insights to proposal generators used in other object detection/instance segmentation tasks.

\clearpage
%
%
\bibliographystyle{splncs04}
\bibliography{egbib}

\begin{thebibliography}{10}
\providecommand{\url}[1]{\texttt{#1}}
\providecommand{\urlprefix}{URL }
\providecommand{\doi}[1]{https://doi.org/#1}

\bibitem{craft}
Baek, Y., Lee, B., Han, D., Yun, S., Lee, H.: Character region awareness for
  text detection. In: Proc. Conf. Comput. Vision Pattern Recognition. pp.
  9365--9374 (2019)

\bibitem{bissacco2013photoocr}
Bissacco, A., Cummins, M., Netzer, Y., Neven, H.: {PhotoOCR}: Reading text in
  uncontrolled conditions. In: Proc. Int. Conf. Comput. Vision. pp. 785--792
  (2013)

\bibitem{Busta_2017_ICCV}
Busta, M., Neumann, L., Matas, J.: {Deep TextSpotter}: An end-to-end trainable
  scene text localization and recognition framework. In: Proc. Int. Conf.
  Comput. Vision. pp. 2223--2231 (2017)

\bibitem{total-text}
Ch'ng, C.K., Chan, C.S.: Total-text: A comprehensive dataset for scene text
  detection and recognition. In: Proc. Int. Conf. on Document Analysis and
  Recognition. pp. 935--942 (2017)

\bibitem{CK2019}
Ch'ng, C.K., Chan, C.S., Liu, C.: Total-text: Towards orientation robustness in
  scene text detection. Int. J. on Document Analysis and Recognition  (2019)

\bibitem{deng2018pixellink}
Deng, D., Liu, H., Li, X., Cai, D.: Pixellink: Detecting scene text via
  instance segmentation. In: AAAI Conf. on Artificial Intelligence (2018)

\bibitem{TextDragon}
Feng, W., He, W., Yin, F., Zhang, X.Y., Liu, C.L.: {TextDragon}: An end-to-end
  framework for arbitrary shaped text spotting. In: Proc. Int. Conf. Comput.
  Vision (2019)

\bibitem{fastrcnn}
Girshick, R.B.: Fast {R-CNN}. In: Proc. Int. Conf. Comput. Vision. pp.
  1440--1448 (2015)

\bibitem{gupta2016synthetic}
Gupta, A., Vedaldi, A., Zisserman, A.: Synthetic data for text localisation in
  natural images. In: Proc. Conf. Comput. Vision Pattern Recognition (2016)

\bibitem{hassner2012computation}
Hassner, T., Rehbein, M., Stokes, P.A., Wolf, L.: Computation and palaeography:
  potentials and limits. Dagstuhl Reports  \textbf{2}(9),  184--199 (2012)

\bibitem{he2017mask}
He, K., Gkioxari, G., Doll{\'a}r, P., Girshick, R.: Mask r-cnn. In: Proc. Int.
  Conf. Comput. Vision. pp. 2961--2969 (2017)

\bibitem{he2016deep}
He, K., Zhang, X., Ren, S., Sun, J.: Deep residual learning for image
  recognition. In: Proc. Conf. Comput. Vision Pattern Recognition. pp. 770--778
  (2016)

\bibitem{he2016accurate}
He, T., Huang, W., Qiao, Y., Yao, J.: Accurate text localization in natural
  image with cascaded convolutional text network. CoRR  \textbf{abs/1603.09423}
  (2016)

\bibitem{he2016text}
He, T., Huang, W., Qiao, Y., Yao, J.: Text-attentional convolutional neural
  network for scene text detection. Trans. Image Processing  \textbf{25}(6),
  2529--2541 (2016)

\bibitem{he2018end}
He, T., Tian, Z., Huang, W., Shen, C., Qiao, Y., Sun, C.: An end-to-end
  textspotter with explicit alignment and attention. In: Proc. Conf. Comput.
  Vision Pattern Recognition. pp. 5020--5029 (2018)

\bibitem{deepdirect}
He, W., Zhang, X., Yin, F., Liu, C.: Deep direct regression for multi-oriented
  scene text detection. In: Proc. Int. Conf. Comput. Vision (2017)

\bibitem{jaderberg2016reading}
Jaderberg, M., Simonyan, K., Vedaldi, A., Zisserman, A.: Reading text in the
  wild with convolutional neural networks. Int. J. Comput. Vision
  \textbf{116}(1),  1--20 (2016)

\bibitem{karatzas2015icdar}
Karatzas, D., Gomez{-}Bigorda, L., Nicolaou, A., Ghosh, S.K., Bagdanov, A.D.,
  Iwamura, M., Matas, J., Neumann, L., Chandrasekhar, V.R., Lu, S., Shafait,
  F., Uchida, S., Valveny, E.: {ICDAR} 2015 competition on robust reading. In:
  Proc. Int. Conf. on Document Analysis and Recognition. pp. 1156--1160 (2015)

\bibitem{karatzas2013icdar}
Karatzas, D., Shafait, F., Uchida, S., Iwamura, M., i~Bigorda, L.G., Mestre,
  S.R., Mas, J., Mota, D.F., Almazan, J.A., de~las Heras, L.P.: {ICDAR} 2013
  robust reading competition. In: Proc. Int. Conf. on Document Analysis and
  Recognition. pp. 1484--1493 (2013)

\bibitem{Li_2017_ICCV}
Li, H., Wang, P., Shen, C.: Towards end-to-end text spotting with convolutional
  recurrent neural networks. In: Proc. Int. Conf. Comput. Vision. pp.
  5248--5256 (2017)

\bibitem{liao2019mask}
Liao, M., Lyu, P., He, M., Yao, C., Wu, W., Bai, X.: {Mask TextSpotter}: An
  end-to-end trainable neural network for spotting text with arbitrary shapes.
  Trans. Pattern Anal. Mach. Intell. pp.~1--1 (2019)

\bibitem{liao2018textboxes++}
Liao, M., Shi, B., Bai, X.: {TextBoxes++}: A single-shot oriented scene text
  detector. Trans. Image Processing  \textbf{27}(8),  3676--3690 (2018)

\bibitem{liao2017textboxes}
Liao, M., Shi, B., Bai, X., Wang, X., Liu, W.: {TextBoxes}: A fast text
  detector with a single deep neural network. In: AAAI Conf. on Artificial
  Intelligence (2017)

\bibitem{LiaoWYCB20}
Liao, M., Wan, Z., Yao, C., Chen, K., Bai, X.: Real-time scene text detection
  with differentiable binarization. In: AAAI Conf. on Artificial Intelligence.
  pp. 11474--11481 (2020)

\bibitem{liao2018rotation}
Liao, M., Zhu, Z., Shi, B., Xia, G.s., Bai, X.: Rotation-sensitive regression
  for oriented scene text detection. In: Proc. Conf. Comput. Vision Pattern
  Recognition. pp. 5909--5918 (2018)

\bibitem{lin2017feature}
Lin, T.Y., Doll{\'a}r, P., Girshick, R., He, K., Hariharan, B., Belongie, S.:
  Feature pyramid networks for object detection. In: Proc. Conf. Comput. Vision
  Pattern Recognition. pp. 2117--2125 (2017)

\bibitem{liu2018fots}
Liu, X., Liang, D., Yan, S., Chen, D., Qiao, Y., Yan, J.: {FOTS}: Fast oriented
  text spotting with a unified network. In: Proc. Conf. Comput. Vision Pattern
  Recognition. pp. 5676--5685 (2018)

\bibitem{abcnet}
Liu, Y., Chen, H., Shen, C., He, T., Jin, L., Wang, L.: Abcnet: Real-time scene
  text spotting with adaptive bezier-curve network. In: Proc. Conf. Comput.
  Vision Pattern Recognition. pp. 9809--9818 (2020)

\bibitem{mcn}
Liu, Z., Lin, G., Yang, S., Feng, J., Lin, W., Goh, W.L.: Learning markov
  clustering networks for scene text detection. In: Proc. Conf. Comput. Vision
  Pattern Recognition. pp. 6936--6944 (2018)

\bibitem{LyuLYWB18}
Lyu, P., Liao, M., Yao, C., Wu, W., Bai, X.: {Mask TextSpotter}: An end-to-end
  trainable neural network for spotting text with arbitrary shapes. In:
  European Conf. Comput. Vision. pp. 71--88 (2018)

\bibitem{lyu2018multi}
Lyu, P., Yao, C., Wu, W., Yan, S., Bai, X.: Multi-oriented scene text detection
  via corner localization and region segmentation. In: Proc. Conf. Comput.
  Vision Pattern Recognition. pp. 7553--7563 (2018)

\bibitem{milletari2016v}
Milletari, F., Navab, N., Ahmadi, S.A.: V-net: Fully convolutional neural
  networks for volumetric medical image segmentation. In: Int. Conf. on 3D
  Vision. pp. 565--571 (2016)

\bibitem{hourglass}
Newell, A., Yang, K., Deng, J.: Stacked hourglass networks for human pose
  estimation. In: European Conf. Comput. Vision. pp. 483--499 (2016)

\bibitem{qin2019towards}
Qin, S., Bissacco, A., Raptis, M., Fujii, Y., Xiao, Y.: Towards unconstrained
  end-to-end text spotting. In: Proc. Int. Conf. Comput. Vision (2019)

\bibitem{ren2015faster}
Ren, S., He, K., Girshick, R., Sun, J.: Faster r-cnn: Towards real-time object
  detection with region proposal networks. In: Neural Inform. Process. Syst.
  pp. 91--99 (2015)

\bibitem{ronneberger2015u}
Ronneberger, O., Fischer, P., Brox, T.: U-net: Convolutional networks for
  biomedical image segmentation. In: Int. Conf. on Medical image computing and
  computer-assisted intervention. pp. 234--241. Springer (2015)

\bibitem{shi2016end}
Shi, B., Bai, X., Yao, C.: An end-to-end trainable neural network for
  image-based sequence recognition and its application to scene text
  recognition. Trans. Pattern Anal. Mach. Intell.  \textbf{39}(11),  2298--2304
  (2017)

\bibitem{tian2019learning}
Tian, Z., Shu, M., Lyu, P., Li, R., Zhou, C., Shen, X., Jia, J.: Learning
  shape-aware embedding for scene text detection. In: Proc. Conf. Comput.
  Vision Pattern Recognition. pp. 4234--4243 (2019)

\bibitem{vati1992generic}
Vatti, B.R.: A generic solution to polygon clipping. Communications of the ACM
  \textbf{35}(7),  56--64 (1992)

\bibitem{WangLZYBXHW020}
Wang, H., Lu, P., Zhang, H., Yang, M., Bai, X., Xu, Y., He, M., Wang, Y., Liu,
  W.: All you need is boundary: Toward arbitrary-shaped text spotting. In: AAAI
  Conf. on Artificial Intelligence. pp. 12160--12167 (2020)

\bibitem{WangWCN12}
Wang, T., Wu, D.J., Coates, A., Ng, A.Y.: End-to-end text recognition with
  convolutional neural networks. In: Int. Conf. Pattern Recognition (2012)

\bibitem{wang2019shape}
Wang, W., Xie, E., Li, X., Hou, W., Lu, T., Yu, G., Shao, S.: Shape robust text
  detection with progressive scale expansion network. In: Proc. Conf. Comput.
  Vision Pattern Recognition. pp. 9336--9345 (2019)

\bibitem{xing2019charnet}
Xing, L., Tian, Z., Huang, W., Scott, M.R.: Convolutional character networks.
  In: Proc. Int. Conf. Comput. Vision (2019)

\bibitem{xue2018accurate}
Xue, C., Lu, S., Zhan, F.: Accurate scene text detection through border
  semantics awareness and bootstrapping. In: European Conf. Comput. Vision. pp.
  355--372 (2018)

\bibitem{MSRA}
Yao, C., Bai, X., Liu, W., Ma, Y., Tu, Z.: Detecting texts of arbitrary
  orientations in natural images. In: Proc. Conf. Comput. Vision Pattern
  Recognition (2012)

\bibitem{Zhan_2019_ICCV}
Zhan, F., Xue, C., Lu, S.: Ga-dan: Geometry-aware domain adaptation network for
  scene text detection and recognition. In: Proc. Int. Conf. Comput. Vision
  (2019)

\bibitem{zhang2016multi}
Zhang, Z., Zhang, C., Shen, W., Yao, C., Liu, W., Bai, X.: Multi-oriented text
  detection with fully convolutional networks. In: Proc. Conf. Comput. Vision
  Pattern Recognition. pp. 4159--4167 (2016)

\bibitem{zhong2016deeptext}
Zhong, Z., Jin, L., Zhang, S., Feng, Z.: {DeepText}: {A} unified framework for
  text proposal generation and text detection in natural images. CoRR
  \textbf{abs/1605.07314} (2016)

\bibitem{zhou2017east}
Zhou, X., Yao, C., Wen, H., Wang, Y., Zhou, S., He, W., Liang, J.: {EAST:} an
  efficient and accurate scene text detector. In: Proc. Conf. Comput. Vision
  Pattern Recognition. pp. 2642--2651 (2017)

\end{thebibliography}

\clearpage
\appendix
\section{Methodology details}

\begin{table}[ht]
    \setlength{\tabcolsep}{8.0pt}
    \centering
    \caption{\textbf{Illustration of our SPN segmentation prediction module.} ``Conv'': convolution operator; ``BN'': batch normalization; ``DeConv'': de-convolution operator. ``k'', ``s'', and ``p'' are short for kernel size, stride, and padding respectively.}
    \begin{tabularx}{0.65\linewidth}{@{}lcc@{}}
    \toprule
    Tpye    & Configurations           & Input/output channel \\ 
    \midrule
    Conv    & k: 3; s: 1; p: 1         & 256/64               \\ 
    BN      & momentum: 0.1 & 64/64                \\ 
    ReLU    & -                        & 64/64                \\ 
    DeConv  & k: 2, s: 2, p: 0         & 64/64                \\ 
    BN      & momentum: 0.1 & 64/64                \\ 
    ReLU    & -                        & 64/64                \\ 
    DeConv  & k: 2, s: 2, p: 0         & 64/1                 \\ 
    Sigmoid & -                        & 1/1                  \\ 
    \bottomrule
    \end{tabularx}
    \label{tab:predict}
\end{table}

\section{Rotation robustness}
\begin{figure*}[ht]
    \centering
    \includegraphics[width=1.0\linewidth]{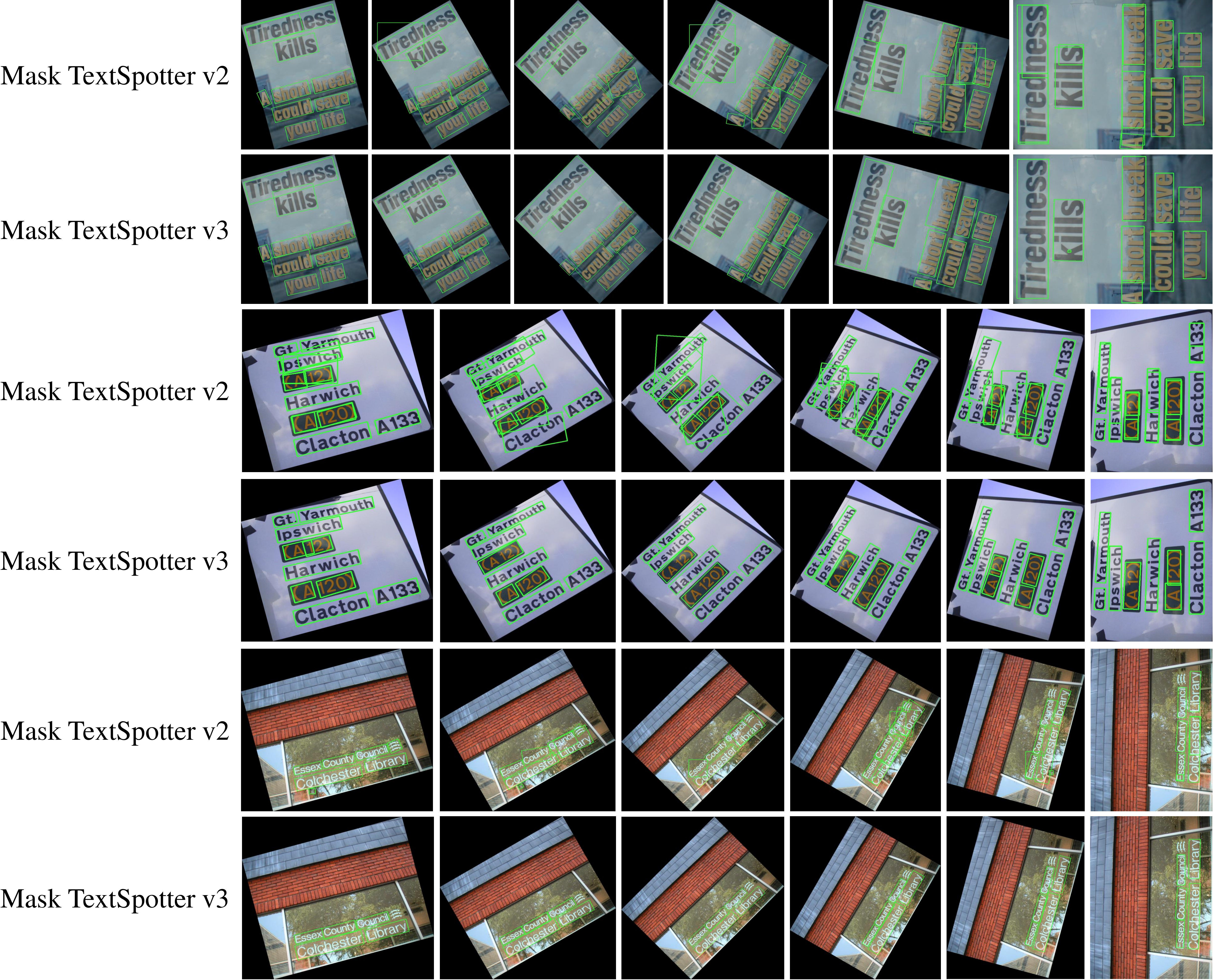}
    \caption{\textbf{Qualitative results on the Rotated ICDAR 2013 dataset.} The rotating angles are $15^\circ$, $30^\circ$, $45^\circ$, $60^\circ$ , $75^\circ$, and $90^\circ$ for the columns from left to right.}
    \label{fig:supp_visu}
\end{figure*}

More qualitative amd quantitative results on the Rotated ICDAR 2013 dataset are shown in Fig.~\ref{fig:supp_visu}, Tab.~\ref{tab:rotated_ic13_det}, and Tab.~\ref{tab:rotated_ic13_rec}.

\begin{table*}[ht]
    \setlength{\tabcolsep}{8.0pt}
    \label{tab:rotated_ic13_det}
    \centering
    \caption{\textbf{Quantitative detection results on the Rotated ICDAR 2013 dataset.} The evaluation protocol is the same as the one in ICDAR 2015 dataset. *CharNet is tested with the official released pre-trained model; Mask TextSpotter v2 is trained with the same rotating augmentation as Mask TextSpotter v3. ``RA'' is short for rotating angles. ``P'', ``R'', and ``F'' indicate precision, recall and F-measure respectively.}
    \begin{tabularx}{1.0\textwidth}{c*{9}c}
    \toprule
    \multirow{2}{*}{RA ($^\circ$)} & \multicolumn{3}{c}{CharNet} & \multicolumn{3}{c}{Mask TextSpotter v2} & \multicolumn{3}{c}{Mask TextSpotter v3} \\ \cline{2-10} 
                                     & P       & R       & F       & P           & R           & F           & P          & R          & F         \\
    \midrule
    0                               & 82.3    & 81.7    & 82      & 89.9        & 85.3        & 87.5        & 90.5       & 84.4       & 87.4      \\
    15                              & 88.1    & 82.2    & 85.1    & 84.6        & 77.4        & 80.7        & 91.8       & 82.3       & 86.8      \\
    30                              & 85.7    & 79.4    & 82.5    & 75.2        & 66          & 70.3        & 91.3       & 78.9       & 84.6      \\
    45                              & 57.8    & 56.6    & 57.2    & 64.8        & 59.9        & 62.2        & 91.6       & 77.9       & 84.2      \\
    60                              & 65.5    & 53.3    & 58.8    & 70.5        & 61.2        & 65.5        & 90.7       & 79.4       & 84.7      \\
    75                              & 58.4    & 41.1    & 48.3    & 77.1        & 77.7        & 77.4        & 89.3       & 80.8       & 84.8      \\
    90                              & 63.0      & 40.4    & 49.2    & 89.8        & 76.8        & 82.8        & 89.8       & 77.2       & 83.0  \\
    \bottomrule  
    \end{tabularx}
\end{table*}

\begin{table*}[ht]
    \setlength{\tabcolsep}{8.0pt}
    \centering
    \caption{\textbf{Quantitative end-to-end recognition results (without lexicon) on the Rotated ICDAR 2013 dataset.} The evaluation protocol is the same as the one in ICDAR 2015 dataset. *CharNet is tested with the official released pre-trained model; Mask TextSpotter v2 is trained with the same rotating augmentation as Mask TextSpotter v3. ``RA'' is short for rotating angles. ``P'', ``R'', and ``F'' indicate precision, recall and F-measure respectively.}
    \label{tab:rotated_ic13_rec}
    \begin{tabularx}{1.0\textwidth}{c*{9}c}
    \toprule
    \multirow{2}{*}{RA ($^\circ$)} & \multicolumn{3}{c}{CharNet} & \multicolumn{3}{c}{Mask TextSpotter v2} & \multicolumn{3}{c}{Mask TextSpotter v3} \\ \cline{2-10} 
                                     & P        & R       & F       & P            & R           & F           & P          & R          & F          \\ 
    \midrule
    0                                & 61.7     & 61.2    & 61.4    & 86.3         & 75.2        & 80.3        & 89.0         & 73         & 80.2       \\ 
    15                               & 66.3     & 61.9    & 64      & 78.4         & 53.5        & 63.6        & 87.2       & 69.8       & 77.5       \\ 
    30                               & 60.9     & 56.5    & 58.6    & 73.9         & 54.7        & 62.9        & 87.8       & 67.5       & 76.3       \\ 
    45                               & 34.2     & 33.5    & 33.9    & 66.4         & 45.8        & 54.2        & 88.5       & 66.8       & 76.1       \\ 
    60                               & 10.3     & 8.4     & 9.3     & 68.2         & 48.3        & 56.6        & 88.5       & 67.6       & 76.6       \\ 
    75                               & 0.3      & 0.2     & 0.2     & 77.0         & 59.2        & 67.0        & 86.9       & 67.6       & 76.0       \\ 
    90                               & 0.0      & 0.0     & 0.0     & 82.0         & 56.9        & 67.1        & 85.9       & 57.9       & 69.1       \\ 
    \bottomrule  
    \end{tabularx}
\end{table*}

\section{Ablation study}
There are two attributions for the RoI masking operator: ``direct/indirect'' and ``soft/hard''. ``direct/indirect'' means using the segmentation/binary map directly or through additional layers; ``soft/hard'' indicates a soft probability mask map whose values are from $[0, 1]$ or a binary polygon mask map whose values are $0$ or $1$. We conduct experiments with the following settings:

\minienumerate{(1) Baseline}: Using the original RoI feature.
\minienumerate{(2) Direct-soft}: It is similar to the RoI masking proposed in Qin et al.~\cite{qin2019towards}, applying element-wise multiplication between the corresponding segmentation probability map and the RoI feature.
\minienumerate{(3) Direct-hard}: Our proposed hard RoI masking, applying element-wise multiplication between the corresponding binary polygon mask map and the RoI feature.
\minienumerate{(4) Indirect-soft}: The corresponding segmentation probability map and the RoI feature are concatenated and then a mask prediction module consisting of two convolutional layers is applied to predict a new mask map. Element-wise multiplication is then applied between the new mask map and RoI feature.
\minienumerate{(5) Indirect-hard}: First, a masked RoI feature is obtained by the hard RoI masking. Then, we concatenate the masked RoI feature and the original RoI feature. Finally, the concatenated feature is classified, choosing whether the masked RoI feature or the original RoI feature is used as the output feature. 

The experimental results in Tab.~\ref{tab_ablation} show that ``direct'' is better than ``indirect'' and ``hard'' is better than ``soft''. The reason is the ``direct'' and ``hard'' strategies provide the most strict mask, fully blocking background noise and neighboring text instances. Our proposed hard RoI masking is simple yet achieves the best performance.

\begin{table}[ht]
    \setlength{\tabcolsep}{8.0pt}
    \centering
    \caption{\textbf{Ablation study on the hard RoI masking.} ``Direct-hard'' indicates our proposed hard RoI masking.}
    \begin{tabularx}{0.55\linewidth}{lccc}
    \toprule
    \multirow{2}{*}{Method}       & \multicolumn{2}{@{}c@{}}{Total-Text} & 
    IC15    \\ \cmidrule{2-4} 
                                  & None           & Full           & Strong        \\ 
    \midrule
    Baseline                & 67.3           & 76.2           & 81.0         \\
    Direct-soft             & 69.1           & 76.0             & 82.2          \\ 
    \textbf{Direct-hard}                   & \textbf{71.2}  & \textbf{78.4}  & \textbf{83.3} \\ 
    Indirect-soft              & 65.8           & 75.6           & 81.2          \\ 
    Indirect-hard & 68.4           & 76.2           & 81.4          \\ 
    \bottomrule
    \end{tabularx}
    \label{tab_ablation}
\end{table}

\end{document}